\documentclass[sigconf]{acmart}

\usepackage{amsmath}
\usepackage{mathtools}
\usepackage{amsthm}
\usepackage{caption}
\usepackage{microtype}
\usepackage{graphicx}
\usepackage{subfigure}
\usepackage{booktabs}
\usepackage{hyperref}
\usepackage{enumitem}
\usepackage{multirow}
\usepackage{multicol}
\usepackage{ulem}
\usepackage{algorithm}
\usepackage{algorithmic}

\usepackage{pifont}
\usepackage{xspace}
\usepackage{scalefnt}
\usepackage{wrapfig}
\usepackage{tabularx} 
\usepackage{colortbl}
\usepackage{xcolor,bm}
\usepackage{bm}

\definecolor{mygrey}{rgb}{0.955,0.95,0.945}

\definecolor{mygreyr}{rgb}{0.99,0.985,0.98}
\definecolor{mygreyr1}{rgb}{0.98,0.975,0.97}
\definecolor{mygreyr2}{rgb}{0.97,0.965,0.96}
\definecolor{mygreyr3}{rgb}{0.96,0.955,0.95}
\definecolor{mygreyr4}{rgb}{0.955,0.95,0.945}
\definecolor{myred}{rgb}{0.995,0.9796,0.958}
\definecolor{myred1}{rgb}{0.992,0.9576,0.932}
\definecolor{myred2}{rgb}{0.991,0.9416,0.917}
\definecolor{myred3}{rgb}{0.990,0.9256,0.902}
\definecolor{myred4}{rgb}{0.990,0.915,0.89}

\AtBeginDocument{%
  }

\setcopyright{acmlicensed}
\copyrightyear{2018}
\acmYear{2018}
\acmDOI{XXXXXXX.XXXXXXX}
\acmConference[Conference acronym 'XX]{Make sure to enter the correct
  conference title from your rights confirmation email}{June 03--05,
  2018}{Woodstock, NY}
\acmISBN{978-1-4503-XXXX-X/2018/06}

\begin{document}

\title{Unlocking Out-of-Distribution Generalization in Dynamics through Physics-Guided Augmentation}
\newcommand{\method}{{\fontfamily{lmtt}\selectfont \textbf{SPARK}}\xspace}

\author{Fan Xu}
\authornote{Fan Xu and Hao Wu are co-first authors. $\dagger$ denotes corresponding authors.}
\affiliation{%
  \institution{University of Science and Technology of China}
  \city{Hefei}
  \state{Anhui}
  \country{China}
}
\email{markxu@mail.ustc.edu.cn}

\author{Hao Wu}
\authornotemark[1]
\affiliation{%
  \institution{Tsinghua University}
  \city{Beijing}
  \country{China}
}
\email{wu-h25@mails.tsinghua.edu.cn}

\author{Kun Wang}
\affiliation{%
  \institution{Nanyang Technological University}
  \country{Singapore}
}
\email{wang.kun@ntu.edu.sg}


\author{Nan Wang}
\affiliation{%
  \institution{Beijing Jiaotong University}
  \city{Beijing}
  \country{China}
}
\email{wangnanbjtu@bjtu.edu.cn}

\author{Qingsong Wen}
\affiliation{%
  \institution{Squirrel AI}
  \city{Seattle}
  \state{Washington}
  \country{USA}
}
\email{qingsongedu@gmail.com}

\author{Xian Wu}
\affiliation{%
  \institution{Tencent}
  \city{Beijing}
  \country{China}
}
\email{kevinxwu@tencent.com}

\author{Wei Gong}
\authornotemark[2]
\affiliation{%
  \institution{University of Science and Technology of China}
  \city{Hefei}
  \state{Anhui}
  \country{China}
}
\email{weigong@ustc.edu.cn}

\author{Xibin Zhao}
\authornotemark[2]
\affiliation{%
  \institution{Tsinghua University}
  \city{Beijing}
  \country{China}
}
\email{zxb@tsinghua.edu.cn}

\renewcommand{\shortauthors}{Fan Xu et al.}


\begin{abstract}
In dynamical system modeling, traditional numerical methods are limited by high computational costs, while modern data-driven approaches struggle with data scarcity and distribution shifts. To address these fundamental limitations, we first propose \method{}, a physics-guided quantitative augmentation plugin. Specifically, \method{} utilizes a reconstruction autoencoder to integrate physical parameters into a physics-rich discrete state dictionary. This state dictionary then acts as a structured dictionary of physical states, enabling the creation of new, physically-plausible training samples via principled interpolation in the latent space. Further, for downstream prediction, these augmented representations are seamlessly integrated with a Fourier-enhanced Graph ODE, a combination designed to robustly model the enriched data distribution while capturing long-term temporal dependencies. Extensive experiments on diverse benchmarks demonstrate that \method{} significantly outperforms state-of-the-art baselines, particularly in challenging out-of-distribution scenarios and data-scarce regimes, proving the efficacy of our physics-guided augmentation paradigm.

\end{abstract}




\maketitle

\section{Introduction} \label{Introduction}

Modeling dynamical systems has long been a critical challenge across numerous scientific fields, including fluid dynamics~\citep{li2023nvfi, janny2023eagle, fei2025openck}, molecular dynamics~\citep{brown2013implementing, yang2022learning}, and atmospheric science~\citep{pathak2022fourcastnet, bi2023accurate, wu2025advanced, gao2025neuralom}, et al. 
Conventional numerical methods~\citep{odibat2020numerical}, often rooted in rigorous partial differential equations (PDEs)~\citep{long2018pde, takamoto2022pdebench} and physical theory formula~\citep{lippe2023pde}, offer a robust foundation for modeling the dynamical evolution of complex systems. However, these methods are notoriously limited by their computational cost and sensitivity to different initial conditions and physical parameters.

Recently, numerous data-driven methods leveraging various neural network architectures have been proposed to solve this problem. 
They are committed to design delicate spatial and temporal components~\citep{lippe2023pde, raonic2024convolutional} to capture high-dimensional non-linear dynamical patterns and latent data distribution.
This paper focuses exclusively on scenarios with fixed data observation points. Mainstream approaches select different model architectures based on whether the data is arranged in a regular pattern. Specifically, for irregular grids or complex geometric boundaries, graph neural networks~\citep{kipf2016semi} are employed to capture intricate interactions between nodes and even along the boundaries~\citep{wang2024beno}.

Despite their promising performance, the majority of these data-driven approaches considerably rely on a substantial volume of data and the assumption of distribution invariance~\citep{wang2021respecting, yang2022learning}. 
Formally, this dynamical system modeling task is still highly challenging in:
\ding{172} \textbf{\textit{Lack of Physical Guidance.}}
Some existing methods simply attach an external parameter embedding module~\citep{lakshmikantham2019method, gao2021phygeonet} to the neural networks. However, such methods struggle to capture higher-order correlations between physical priors and data itself, and are difficult to generalize to unseen parameter configurations~\citep{rame2022fishr, wu2024prometheus}.
\ding{173} \textbf{\textit{Data Sparsity.}} 
Data-driven methods tend to require sufficient data. While data acquisition is limited due to the high computational cost of traditional numerical simulations~\citep{schober2019probabilistic} or the practical constraints on sensor usage in several real-world scenarios. 
\ding{174} \textbf{\textit{Out-of-distribution Generalization.}}
Within the dynamical systems, there usually exist two types of distribution shifts, namely environmental distribution shift~\citep{li2022learning, song2023recovering} and temporal distribution shift~\citep{wang2022koopman, lu2024diversify, wu2024pure}.
The former is determined by predefined environmental attributes such as boundary conditions and physical parameters within the dynamical fields, while the latter arises from potential shifts within the data distribution over long-term temporal evolution~\citep{wu2025turb}.

To address these challenges, we propose a universal augmentation plugin, named \method{}, designed to efficiently encode rich physical priors alongside the given observations, enabling compressed representations and facilitating sample augmentation through physics-aware transformations. 
To begin, we extract latent representations of observed samples, where physical parameters are fused through customized positional encodings and customized channel attention mechanisms.
Further, to achieve efficient compression, we utilize a graph neural network framework combined with latent space quantization techniques, constructing a low-cost discretized state dictionary infused with physical priors.
Subsequently, we perform sample selection and query the pre-trained discretized state dictionary for data augmentation, thus mitigating the impact of environmental distribution shift to some extent. 
Finally, to address temporal distribution shifts, we encode historical observations into an initial state through attention mechanism, and then implement a fourier-enhanced graph ODE for effectively long-term prediction.

In summary, this paper makes the following contributions: 
\ding{182} \textbf{\textit{Novel Perspective.}} We are the first to propose a physics-aware compression and augmentation plugin, which highly enhances the generalization capability to diverse physical scenarios.
\ding{183} \textbf{\textit{Modern Architecture.}} In downstream task, we encode historical observations into a latent space based on attention mechanism and introduce a fourier-enhanced graph ODE to overcome temporal distribution shifts and realize efficient long-term predictions.
\ding{184} \textbf{\textit{Verification.}} We demonstrate the generalization and robustness of \method{} under data scarcity and distribution shifts through both experimental evaluations and theoretical analysis.

\section{Related Work}

\subsection{Dynamical System Modeling.} 
Deep neural networks have recently emerged as powerful tools for tackling the challenges associated with dynamics forecasting~\citep{gao2022earthformer, yin2022continuous}, showcasing their capabilities to efficiently model intricate, high-dimensional systems. To handle complex spatiotemporal dependencies, various advanced models based on convolutional neural networks (CNN)~\citep{ren2022phycrnet, raonic2024convolutional}, recurrent neural networks (RNN)~\citep{mohajerin2019multistep, maulik2021reduced}, Transformer~\citep{wu2023solving, chen2024multi}, or exquisite hybrid architectures~\citep{shi2015convolutional, wu2024pastnet} have been proposed. 
Recently, Neural Operators~\citep{li2021fourier, rahman2022u, anonymous2023factorized} become a popular data-driven approach by learning to approximate the past-future infinite-dimensional function space mappings.
Additionally, Physics-Informed Neural Networks (PINN)~\citep{Raissi2019PhysicsinformedNN, Wang2020WhenAW} integrate prior physical knowledge as additional regularizers into the training process for physical constraints, particularly in systems governed by partial differential equations.
Moreover, most of these approaches are limited to structured grids and lack the ability to handle irregular grids or varying connectivity. Thus researchers have turned to graph neural networks~\citep{fan2019graph, gao2025oneforecast, xu2025breaking} as a promising alternative to accommodate a broader range of scenarios. 
GNNs inherit key physical properties from geometric deep learning, including permutation invariance and spatial equivariance~\citep{Li2020NeuralOG, wu2022learning}, which offer distinct advantages for modeling dynamical systems.

\subsection{Out-of-distribution Generalization.} 
Out-of-distribution (OOD) generalization~\citep{liu2021towards, hendrycks2021many, deng2023counterfactual} has emerged as a critical challenge in machine learning, especially for models that encounter distribution shifts between training and testing data.
Significant advancements have been made in OOD generalization techniques, containing invariant causal inference~\citep{gui2023joint}, data augmentation~\citep{wang2024nuwadynamics}, domain adaptation~\citep{kundu2020towards, garg2022domain}, and adversarial training~\citep{deng2023counterfactual, zhan2023pareto} et al.
However, most existing work focuses on static scenarios~\citep{gui2023joint}. Unlike in static settings, distribution shifts in dynamics are often not arbitrary but are rooted in changes to the underlying physical laws. 
These dynamic systems exhibit more complicated distribution patterns corresponding to varying system properties or spatiotemporal environments, which is the primary focus of our study.
In this work, we propose \method{}, which compresses rich physical information into a discrete memory bank for generalization, and design a fourier-enhanced graph ODE to relieve temporal distribution shifts.

\section{Methodology}

\begin{figure*}[!t]
\centering
\includegraphics[width=1.0\linewidth]{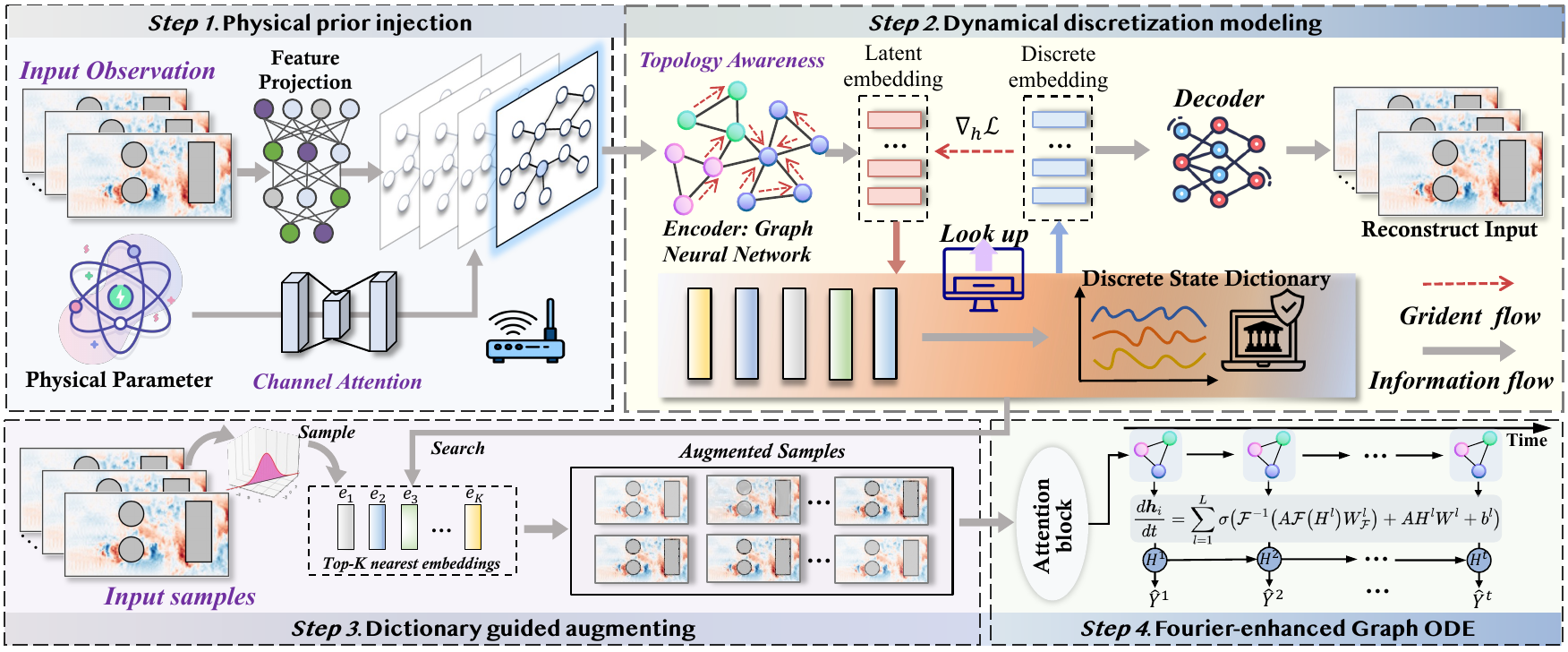}
\vspace{-10pt}
\caption{\textbf{Model overview} -- The proposed \method{} consists of four steps: 
\ding{182} Physical prior incorporation, where physical parameters are encoded into the input observations; 
\ding{183} Dynamical discretization modeling through reconstruction to create a discrete physics-rich state dictionary; 
\ding{184} State dictionary guided augmenting on sampled training data; and 
\ding{185} Fourier-enhanced graph ODE for dynamical system prediction based on historical observations.}
\label{fig:frame}
\end{figure*}

\subsection{Problem Definition}

Given a dynamical system governed by physical laws such as PDEs, we aim to enhance prediction using autoencoder reconstruction and discrete quantization. We have $N$ observation points in the spatial domain $\Omega$, located at $s = \{s_1, \cdots, s_N\}$, where $s_i \in \mathbb{R}^{d_s}$. At time step $t$, the observations are $\mathcal{X}^t = \{{x}_1^t, \cdots, {x}_N^t\}$, where ${x}_i^t \in \mathbb{R}^{d}$ and $d$ represents the number of observation channels. Physical parameters affect the dynamical system, leading to different conditions and distribution shifts. We first employ reconstruction model and construct a discrete state dictionary to compress and store physical prior information. Then, given historical observation sequences $\{\mathcal{X}_i^{-T_0+1:0}\}_{i=1}^N$, our goal is to use the pre-trained state dictionary for data augmentation and predict future observations $\{\mathcal{Y}_i^{1:T}\}_{i=1}^N$ at each observation point.

\subsection{Framework Overview}

In this section, we systematically introduce our \method{} framework, as shown in Figure~\ref{fig:frame}.
We first introduce a coupled reconstruction autoencoder, which incorporates physical parameters, to generate a discrete, physics-rich state dictionary for compression.
Subsequently, in downstream task, we employ the pre-trained state dictionary to augment the training samples based on the fusion mechanism.
We then utilize the attention mechanism to model historical observations and propose a fourier-based graph ODE to realize downstream dynamical system prediction.

\subsection{Physics-incorporated Data Compression}
Here, we seek to transform the dynamical system observations, along with rich physical parameters, into a compressed structured representation. We then maintain a discrete state dictionary and leverage graph neural networks to reconstruct the observations based on this representation.

\paragraph{Physical Parameter Guided Channel Attention.}
Inspired by~\citep{takamoto2023learning}, we employ a channel attention mechanism to effectively embed external parameters into the neural networks, facilitating the transfer of parameter information into the latent space. 
Hereafter, we refer to the neural network parameters as "weights" to avoid terminological conflicts with the physical parameters in dynamical systems. 
Specifically, the channel attention obtains two $d$-dimentional mask attention vectors $a_{\omega} \in \mathbb{R}^{d}~(\omega = 1,2)$ from the parameters $\delta$ using a 2-layer MLP. 
\begin{equation}
a_\omega=W_{2,\omega} \cdot \sigma(W_{1,\omega}\delta+b_1)+b_2,
\end{equation}
where $W_{1/2,\omega}$ is the weight matrix, $b_{1/2}$ is the bias term, and $\sigma$ is the \text{GeLU} activation function. 
For the node feature $x$, we employ two convolutional opertors: a $1 \times 1$ convolution ($g_1$) and a spectral convolution ($g_2$). 
The former one ensures fine-grained alignment between channels, and the latter one operates at a global frequency-domain and captures the broader and structural correlations.
Then the obtained representations are multiplied by the mask attention vectors separately and finally combined to realize the channel attention between observations and parameters. 
\begin{equation}
    \tilde{x}_{i, \omega}=g_\omega(x_i), ~ 
    h_{i, \omega}=a_{\omega} \odot \tilde{x}_{i, \omega} ~
    \Rightarrow ~ h_i = x_i + h_{i,1} + h_{i,2},
\end{equation}
where $\odot$ is the Hadamard operator. By this way, we achieve parameter channel fusion for physical processes through attention-enhanced convolutional networks.

\paragraph{Physics-Embedded Reconstruction with Discrete State Dictionary.} 
Now we have obtained the observation features $h$ fusing with the physical parameter information. 
For reconstruction, we propose to tokenize each node as discrete embeddings using a $L$-layer GNN encoder~\citep{xu2024revisiting} and a variant of \textit{VQ-VAE}~\citep{van2017neural, wang2025beamvq}. 
Formally, in the $l$-th layer of the GNN encoder, the output representations $h^{(l)}$ are combined from the previous representations using aggregation operator and residual function as: 
\begin{equation} \label{boun}
\begin{aligned}
h_i^{'(l)}&=\text{AGGREGATE}\left(\left\{h_j^{(l-1)}:j\in\mathcal{N}(i)\right\}\right), \\
h_i^{(l)}&=\text{COMBINE}\left(h_i^{(l-1)},h_i^{'(l)}\right).
\end{aligned}
\end{equation}
The final output of the encoder, $h_i^{(L)}$, is a high-level latent representation of the node's state.
Then, the selector look up the nearest neighbor code embedding $z_i$ in the maintained state dictionary $E=\left[e_1, e_2, \cdots, e_M\right] \in \mathbb{R}^{M \times D}$ ($M$ denotes the state dictionary size) for each node embedding $h_i$. 
\begin{equation}\label{eq5}
    \breve{h}_i = e_{k} \quad \text{where} \quad k = \operatorname{argmin}_j \|h^{(L)}_i - e_j\|_2^2.
\end{equation}
Finally, the discretized representations $\breve{h}_i$ are fed into a 2-layer \text{MLP} decoder $\mathcal{I}(\cdot)$ to reconstruct the input observations for an end-to-end optimization.

\paragraph{Pre-training Loss Function.}
For the whole reconstruction pre-training, we minimize the reconstruction loss and the discrete state dictionary loss simultaneously. Specifically, to overcome the training challenges associated with discretization encoding, we utilize the stopgradient operator $\text{sg}(\cdot)$~\citep{van2017neural}. 
The whole pre-training loss function $\mathcal{L}_{pre}$ is as follows:
\begin{equation}
\begin{aligned}
\mathcal{L}_{pre} &= \frac{1}{T N} \sum_{t=1}^T \sum_{i=1}^N\left(\hat{\mathcal{X}}_{i}^{t}-\mathcal{X}_{i}^{t}\right)^2  \\
&+ \frac{1}{T N} \sum_{t=1}^T \sum_{i=1}^N \left (\mu \left\|h_{i}^{t}-\text{sg}[e]\right\|_2^2+\gamma\left\|\text{sg}\left[h_{i}^{t}\right]-e\right\|_2^2\right ),
\end{aligned}
\end{equation}
where $\hat{\mathcal{X}_{i}^{t}}$ and $\mathcal{X}_{i}^{t}$ denote the reconstructed and the initial node embedding, respectively. And $\mu$, $\gamma$ are the hyperparameters to balance these loss components.

\subsection{Downstream Data Augmentation and Dynamical System Prediction Training}

\paragraph{Augmentation via Physical State Dictionary.}
With the physics-rich discrete state dictionary $E$ pre-trained, we now detail how it is employed to guide the data augmentation. Each codebook vector $e_j \in E$ represents a learned, physically-valid prototype, and together they partition the continuous latent space into a finite set of representative physical conditions. 

Instead of using the continuous latent vector $h_i$ directly, we perform vector quantization by replacing it with its single nearest prototype, $\breve{h}_i$, from the dictionary (according to Eq.~\ref{eq5}).
This act of "snapping" to the closest physical prototype serves as a powerful form of \textit{representation augmentation}; it filters out instance-specific noise and forces the model to operate on a robust, physically-grounded state space. 
Consequently, the downstream dynamics model is trained not on raw latent observations, but on a \textit{canonicalized sequence of physical states}. 
This principled interpolation enhances physical diversity and smooths the latent space, preventing overfitting to a single prototype. 
The resulting augmented samples are then incorporated into the training pipeline on-the-fly to enhance the model's generalization performance.

\paragraph{Fourier-enhanced Graph ODE for Prediction.}
In downstream dynamical system prediction, we first manage to map the historical observations of individual node into a latent representation. 
Specifically, we employ an attention mechanism to calculate the corresponding attention score of each time step, and then initialize the observation state vector based on this. In formulation:
\begin{equation}
h_{i}=\frac1{T_0}\sum_{t=1}^{T_0}\delta(\alpha_{i}^{t} 
\cdot h_{i}^{t}), ~~~~ \alpha_{i}^{t}= \left(h_{i}^{t} \right )^{T} \mathrm{tanh}\left(\left(\frac{1}{T_0} \sum_{t=1}^{T_0} h_{i}^{t}\right)W_{\alpha}\right),
\end{equation}
where $\alpha_{i}^{t}$ denotes the attention score of node $i$ at time step $t$, and $W_{\alpha}$ represents the trainable parameter.
We further propose a fourier-enhanced graph ODE to model the obtained state vector by integration of the spectral global feature and the local feature. In formulation:
\begin{equation}
\begin{aligned}
\frac{d h_{i}}{dt} &= \Phi \left(h_{1}, h_{2} \cdot \cdot \cdot h_{N}, \mathcal{G},\Theta \right)  \\
&= \sum_{l=1}^{L} \sigma \left( \mathcal{F}^{-1} \left ( A\mathcal{F}\left( H^l\right)W_{\mathcal{F}}^l \right) + AH^lW^l + b^l\right).
\end{aligned}
\end{equation}
Here, $\mathcal{G}$ and $\Theta$ are the graph structure and the whole model parameters, $\mathcal{F}$ denotes \textit{Discrete Fourier Transform (DFT)}~\citep{sundararajan2001discrete}, $A$ is the adjacency matrix, and $W_{\mathcal{F}}$, $W$, $b$ are the matrix weights.
Then, the latent dynamics can be solved by any ODE solver like Runge-Kutta~\citep{schober2014probabilistic}.
\begin{equation}
\begin{aligned}
h_i^{t_1} \cdots h_i^{t_T} &= \mathrm{ODESolve}(\Phi,[h_1^{0},h_2^{0}\cdots h_N^{0}])  \\
&\Rightarrow ~ \widehat{y}_i^t|h_i^t\sim p(\widehat{y}_i^t|f_{\mathrm{dec}}(h_i^t)),
\end{aligned}
\end{equation}
where $\{h_i^{t_1}\cdots h_i^{t_T} \}$ represents the latent future vector of node $i$, and $\widehat{y}_i^{t}$ is the corresponding predicted vector through a stacked two-layer MLP decoder $f_{dec}(\cdot)$.

\paragraph{Training.}
The training objective for the dynamics system prediction is to learn a mapping from the input observations $\mathcal{X}$ to future states $\mathcal{Y}$, where the system evolves according to certain physical laws like parameters. Formally, we aim to minimize the following loss function:
\begin{equation}
    \mathcal{L}_{\text{dyn}} = \frac{1}{TN} \sum_{i=1}^T \sum_{i=1}^N \|\hat{\mathcal{Y}}_{i}^{t} - \mathcal{Y}_{i}^{t}\|_2^2 + \lambda_{\text{reg}} \mathcal{R}(\theta),
\end{equation}
where $\hat{\mathcal{Y}}$ denotes the predicted future states, and $\mathcal{R}(\theta)$ is a regularization term on the model parameters $\theta$ to prevent overfitting. 
Finally, to seamlessly integrate the augmented samples into the training pipeline, we adopt a curriculum learning approach, gradually adding the augmented samples into the training set. Initially, the model is trained on the original dataset, and as training progresses, the proportion of augmented samples is increased, allowing the model to adapt to a richer variety of dynamical behaviors.

\subsection{Theoretical Analysis}

\begin{theorem}[\textit{\textbf{Enhancement of Model Generalization via Physical Priors from an Information-Theoretic Perspective}}]\label{theorem1}

Let $\mathcal{D}$ be the training dataset, $\theta$ be the model parameters, and $\mathcal{P}$ be the physical prior information. Assume that the conditional mutual information between $\theta$ and $\mathcal{D}$ given $\mathcal{P}$ is $I(\theta; \mathcal{D} \mid \mathcal{P})$.
For an i.i.d. training dataset of size $n$, the upper bound on the expected generalization error is:
\begin{equation}
\left| \mathbb{E}_{\theta, \mathcal{D}}\left[ \mathcal{L}(\theta) - \mathcal{L}_{\text{emp}}(\theta; \mathcal{D}) \right] \right| \leq \sqrt{\dfrac{2 I(\theta; \mathcal{D} \mid \mathcal{P})}{n}},
\end{equation}
where $\mathcal{L}(\theta)$ is the expected loss under the true distribution, and $\mathcal{L}_{\text{emp}}(\theta; \mathcal{D})$ is the empirical loss on the training data.
\end{theorem}

Thus, introducing physical prior information $\mathcal{P}$ reduces the conditional mutual information $I(\theta; \mathcal{D} \mid \mathcal{P})$, which, by the above theorem, decreases the upper bound of the generalization error. This implies that physical priors enhance model generalization, improving performance on unseen data. The detailed theorem with the proof is illustrated in Appendix~\ref{sec:theo1}.

\begin{theorem}[\textit{\textbf{Upper Bound on Generalization Error in Bayesian Learning with Physical Priors}}]\label{theorem2}

Let $\mathcal{H}$ be a hypothesis space, $\theta \in \mathcal{H}$ be the model parameters, and $\mathcal{D} = \{(\mathcal{X}_i, \mathcal{Y}_i)\}_{i=1}^N$ be the training dataset. Let $\ell(\theta; \mathcal{X}, \mathcal{Y})$ be the loss function, with the true risk defined as $\mathcal{L}(\theta) = \mathbb{E}_{(\mathcal{X}, \mathcal{Y}) \sim \mathcal{P}_{\text{data}}}[\ell(\theta; \mathcal{X}, \mathcal{Y})]$ and the empirical risk as $\mathcal{L}_{\text{emp}}(\theta) = \frac{1}{N} \sum_{i=1}^N \ell(\theta; \mathcal{X}_i, \mathcal{Y}_i)$.

Assume the prior distribution $P(\theta)$ incorporates physical prior information, and the posterior distribution is $Q(\theta)$. For any $\delta > 0$, with probability at least $1 - \delta$, the following upper bound on the generalization error holds:
\begin{equation}
\mathbb{E}_{\theta \sim Q} [\mathcal{L}(\theta)] \leq \mathbb{E}_{\theta \sim Q} [\mathcal{L}_{\text{emp}}(\theta)] + \sqrt{\dfrac{KL(Q \| P) + \ln \dfrac{2\sqrt{N}}{\delta}}{2N}},
\end{equation}
where $KL(Q \| P)$ is the Kullback-Leibler divergence~\citep{van2014renyi} between the posterior distribution $Q$ and the prior distribution $P$.
\end{theorem}

By incorporating physical prior information as the prior $P(\theta)$, the KL divergence $KL(Q \| P)$ between the posterior $Q(\theta)$ and the prior $P(\theta)$ is reduced, thereby lowering the upper bound on the generalization error. This demonstrates that physical priors enhance model generalization in the Bayesian framework. 
The detailed proof can be seen in Appendix~\ref{sec:theo2}.
\begin{table*}[!t] 
\tiny
    \begin{sc}
    \centering
    \caption{Detailed setup for the \textit{In-Domain} and \textit{Out-Domain} environments of various benchmarks.}
    \vspace{-10pt}
    \resizebox{0.88\textwidth}{!}{
    \begin{small}        
    \begin{tabular}{c|c|c}
    \toprule
        Benchmarks & In-Domain environments & Adaptation environments \\ 
        \midrule
        Prometheus & $\{a_1, a_2, ..., a_{25}\}$, $\{b_1, b_2, ..., b_{20}\}$ & $\{a_{26}, a_{27}, ..., a_{30}\}$,  $\{b_{21}, b_{22}, ..., b_{25}\}$ \\ 
        2D Navier-Stokes equation & $\nu = \{1e^{-1}, 1e^{-2}, ..., 1e^{-7}, 1e^{-8}\}$ & $\nu = \{1e^{-9}, 1e^{-10}\}$ \\ 
        Spherical Shallow Water equation & $\nu = \{1e^{-1}, 1e^{-2}, ..., 1e^{-7}, 1e^{-8}\}$ & $\nu_t = \{1e^{-9}, 1e^{-10}\}$ \\ 
        3D Reaction–Diffusion equations & $D = \{2.1 \times 10^{-5}, 1.6 \times 10^{-5}, 6.1 \times 10^{-5}\}$ & $D = \{2.03 \times 10^{-9}, 1.96 \times 10^{-9}\}$ \\ 
        ERA5 & $V = \{Sp, SST, SSH, T2m\}$ & $V = \{SSR, SSS\}$ \\ 
        \bottomrule
    \end{tabular} \label{tab:ood_data}
    \end{small}}
    \end{sc}
    \vspace{-5pt}
\end{table*}

\section{Experiment} \label{experiment}

In this section, we present experimental results to demonstrate the effectiveness of the \method{} framework. Our experiments are designed to address the following research questions: 
\begin{itemize}
    \item[$\mathcal{RQ}$1:] Does \method{} effectively handle out-of-distribution generalization while maintaining consistent superiority?
    \item[$\mathcal{RQ}$2:] Can \method{} adapt to more challenging tasks?
    \item[$\mathcal{RQ}$3:] Does \method{} enhance the backbones' generalization?
    \item[$\mathcal{RQ}$4:] Can \method{} maintain scalable and physically consistency?
\end{itemize}

\subsection{Experimental Settings}

\noindent\textbf{Benchmarks.} We choose benchmark datasets from three fields. 
\ding{95} \textit{\textbf{For datasets simulated by industrial software}}, we choose \textit{Prometheus}~\citep{wu2024prometheus} and follow its original environment settings. 
\ding{95} \textit{\textbf{For real-world datasets}}, we select \textit{ERA5}~\citep{hersbach2020era5}, selecting different atmospheric variables like surface pressure (Sp), sea surface temperature (SST), and two-meter temperature (T2m), et al. 
\ding{95} \textit{\textbf{For partial differential equation datasets}}, we examine the \textit{2D Navier-Stokes Equations}~\citep{li2021fourier}, focusing on the effect of viscosity $\nu$ on vorticity, and simulate vorticity under ten different viscosities. We also study the \textit{Spherical Shallow Water Equations}~\citep{galewsky2004initial} to simulate large-scale atmospheric and oceanic flows on Earth's surface, involving viscosity $\nu$, tangential vorticity $w$, and fluid thickness $h$. Additionally, we consider the \textit{3D Reaction-Diffusion Equations}~\citep{rao2023encoding}, describing chemical diffusion and reaction in space, with diffusion coefficient $D$, and involving $u$ and $v$ velocity components. 
More detailed descriptions can be seen in Appendix~\ref{datadescrip}.

\noindent\textbf{Baselines.} 
To comprehensively evaluate the performance of our proposed method, we select two categories of deep learning models as baselines for comparison.
\ding{118} \textit{\textbf{Visual Backbone Networks.}} For visual backbone, we choose ResNet~\citep{he2016deep}, U-Net~\citep{ronneberger2015u}, Vision Transformer (ViT)~\citep{dosovitskiy2021an}, and Swin Transformer (SwinT)~\citep{liu2021swin}. 
\ding{118} \textit{\textbf{Neural Operator Architectures.}} For neural operator, we consider FNO~\citep{li2021fourier}, UNO~\citep{ashiqur2022u}, CNO~\citep{raonic2024convolutional}, and NMO~\citep{wu2024neural}. The details are as follows.

\begin{itemize}[leftmargin=*]
    \item \textbf{U-Net} \cite{ronneberger2015u} is a convolutional neural network originally developed for biomedical image segmentation. Its U-shaped architecture with symmetric skip connections between the encoder and decoder facilitates effective feature integration.
    \item \textbf{ResNet} \cite{he2016deep} introduces residual connections to address the issue of performance degradation in deep networks. These skip connections allow information to bypass layers, enabling deeper and more trainable architectures.
    \item \textbf{ViT} \cite{dosovitskiy2021an} utilizes the Transformer model for image classification. The image is divided into patches, which are processed using self-attention mechanisms, achieving a balance between computational efficiency and accuracy.
    \item \textbf{SwinT} \cite{liu2021swin} employs a sliding window technique for the extraction of both local and global features. This makes it versatile for a wide range of computer vision tasks.
    \item \textbf{FNO} \cite{li2021fourier} leverages Fourier transforms for extracting global features, making it particularly effective for handling continuous field data and solving partial differential equations (PDEs).
    \item \textbf{UNO} \cite{ashiqur2022u} combines U-Net’s architecture with optimization techniques to boost feature extraction and fusion, thereby enhancing the model's overall performance.
    \item \textbf{CNO} \cite{raonic2024convolutional} integrates convolutional operations with operator learning to better handle high-dimensional continuous data, focusing on the modeling of intricate dynamic systems.
    \item \textbf{NMO} \cite{wu2024neural} improves the capability to model dynamical systems by integrating neural networks with manifold learning methods.
\end{itemize}

\noindent\textbf{ID \& OOD Settings.} 
We propose that training and testing in the in-domain parameters is called w/o OOD experiments, while training in the in-domain parameters and testing in the out-domain parameters is called w/ OOD experiments. We present the in-domain and out-domain parameters for different benchmarks in Table~\ref{tab:ood_data}.

\begin{table*}[!t]
    \caption{Comparison of different models on five datasets (Prometheus, Navier–Stokes, Spherical-SWE, 3D Reaction–Diff, ERA5) with and without OOD. Our method achieves the best performance across all benchmarks, especially under OOD conditions. The best-performing result in each column is highlighted in bold, while the second-best is indicated in \underline{underline}.}
    \vspace{-5pt}
    \tiny
    \label{tab:oodmain}
    \centering
    \begin{footnotesize}
        \begin{sc}
            \renewcommand{\multirowsetup}{\centering}
            \setlength{\tabcolsep}{3.5pt} 
            \resizebox{0.98\textwidth}{!}{%
            \begin{tabular}{l|cc|cc|cc|cc|cc}
                \toprule
                \multirow{4}{*}{Model} & \multicolumn{10}{c}{Benchmarks}  \\
                \cmidrule(lr){2-11}
                & \multicolumn{2}{c}{Prometheus} & \multicolumn{2}{c}{Navier–Stokes} & \multicolumn{2}{c}{Spherical-SWE} &  \multicolumn{2}{c}{3D Reaction–Diff} &  \multicolumn{2}{c}{ERA5}  \\
                \cmidrule(lr){2-11}
               & \textit{w/o} OOD & \textit{w/} OOD & \textit{w/o} OOD & \textit{w/} OOD & \textit{w/o} OOD & \textit{w/} OOD & \textit{w/o} OOD & \textit{w/} OOD & \textit{w/o} OOD & \textit{w/} OOD  \\
                \midrule
                U-Net \cite{ronneberger2015u} & 0.0931 & 0.1067 & 0.1982 & 0.2243 & 0.0083 & 0.0087 & 0.0148 & 0.0183 & 0.0843 & 0.0932 \\
                ResNet \cite{he2016deep} & 0.0674 & 0.0696 & 0.1823 & 0.2301 & 0.0081 & 0.0192 & 0.0151 & 0.0186 & 0.0921 & 0.0977 \\
                ViT \cite{dosovitskiy2021an} & 0.0632 & 0.0691 & 0.2342 & 0.2621 & 0.0065 & 0.0072 & 0.0157 & 0.0192 & 0.0762 & 0.0786 \\
                SwinT \cite{liu2021swin} & 0.0652 & 0.0729 & 0.2248 & 0.2554 & 0.0062 & 0.0068 & 0.0155 & 0.0190 & 0.0782 & 0.0832 \\
                \midrule
                FNO \cite{li2021fourier} & 0.0447 & 0.0506 & 0.1556 & 0.1712 & 0.0038 & 0.0045 & 0.0132 & 0.0179 & 0.0723 & 0.0982 \\
                UNO \cite{ashiqur2022u} & 0.0532 & 0.0643 & 0.1764 & 0.1984 & 0.0034 & 0.0041 & 0.0121 & 0.0164 & 0.0665 & 0.0762 \\
                CNO \cite{raonic2024convolutional} & 0.0542 & 0.0655 & 0.1473 & 0.1522 & 0.0037 & 0.0038 & 0.0145 & 0.0182 & 0.0524 & 0.0782 \\
                NMO \cite{wu2024neural} & \underline{0.0397} & \underline{0.0483} & \underline{0.1021} & \underline{0.1032} & \underline{0.0026} & \underline{0.0031} & \underline{0.0129} & \underline{0.0168} & \underline{0.0432} & \underline{0.0563} \\
                \midrule
                \rowcolor{mygrey} Ours & \textbf{0.0294} & \textbf{0.0308} &\textbf{0.0714} & \textbf{0.0772} & \textbf{0.0018} & \textbf{0.0020}& \textbf{0.0102} & \textbf{0.0116} & \textbf{0.0327} & \textbf{0.0344} \\
                \bottomrule
            \end{tabular}
            }
        \end{sc}
    \end{footnotesize}
    \vspace{-5pt}
\end{table*}

\begin{figure*}[!t]
    \centering
    \includegraphics[width=0.98\textwidth]{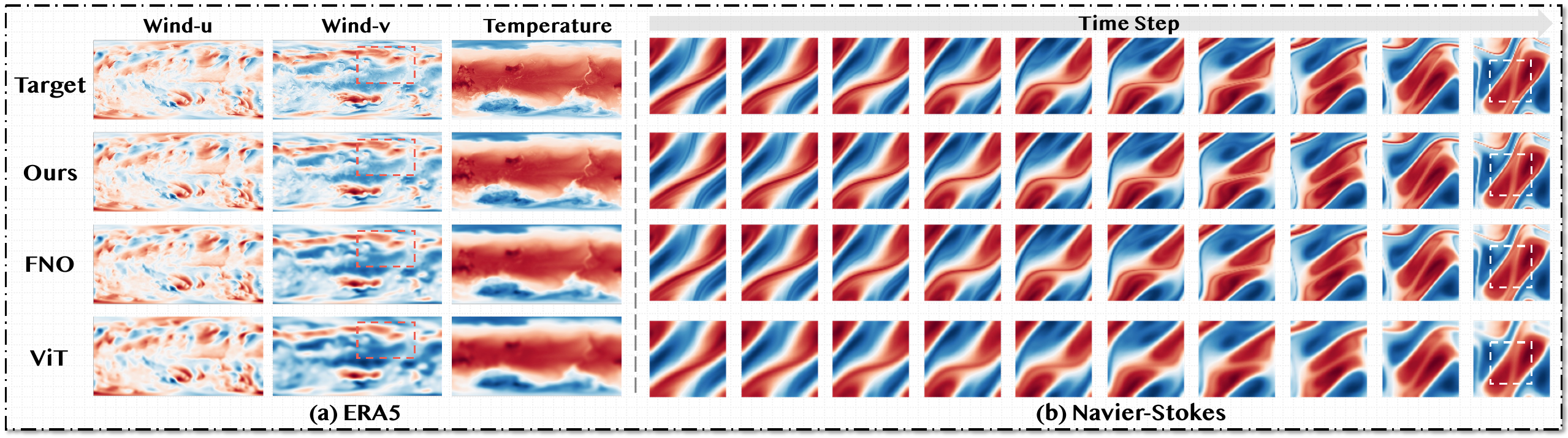}
    \vspace{-5pt}
    \caption{Comparison of Prediction Performance. The figure shows the target values and the predictions from different models (Ours, FNO, ViT) at multiple time steps. It is evident that \method{}'s predictions are closest to the target values, especially in the locally complex regions (highlighted in red or white boxes), demonstrating higher detail-capturing ability and accuracy. }
    \label{fig:vismain} 
\end{figure*}

\noindent\textbf{Metrics Details.} We employ three advanced metrics in our study.

\noindent \ding{224} \textit{\textbf{Mean Squared Error (MSE)}}: This metric provides the average of the squares of the differences between the actual and predicted values. A lower MSE indicates a closer fit of the predictions to the true values. It's given by the equation:
\begin{equation}
\text{MSE} = \frac{1}{N} \sum_{i=1}^{N} (V_{\text{true},i} - V_{\text{fut},i})^2,
\end{equation}
where \( V_{\text{true},i} \) represents the true value, \( V_{\text{fut},i} \) denotes the predicted value, and \( N \) is the number of observations.

\noindent \ding{224} \textit{\textbf{Multi-Scale Structural Similarity (SSIM)}}: SSIM is designed to provide an assessment of the structural integrity and similarity between two images, \( x \) and \( y \). Higher SSIM values suggest that the structures of the two images being compared are more similar.
\begin{equation}
\text{SSIM}(x, y) = \frac{(2\mu_x\mu_y + c_1)(2\sigma_{xy} + c_2)}{(\mu_x^2 + \mu_y^2 + c_1)(\sigma_x^2 + \sigma_y^2 + c_2)},
\end{equation}
where \( \mu \) is the mean, \( \sigma \) represents variance, and \( c_1 \) and \( c_2 \) are constants to avoid instability.

\noindent \ding{224} \textit{\textbf{Peak Signal-to-Noise Ratio (PSNR)}}: PSNR gauges the quality of a reconstructed image compared to its original by measuring the ratio between the maximum possible power of the signal and the power of corrupting noise. A higher PSNR indicates a better reconstruction quality.
\begin{equation}
\text{PSNR} = 10 \times \log_{10} \left( \frac{\text{MAX}_I^2}{\text{MSE}} \right),
\end{equation}
where \( \text{MAX}_I \) is the maximum possible pixel value of the image.

\subsection{Main Results}

As shown in Table~\ref{tab:oodmain}, our method (\method{}) shows better performance on all benchmark datasets, especially with out-of-distribution (OOD) data. For example, in the Prometheus dataset, \method{} achieves an MSE of 0.0294 and 0.0308 for regular and OOD conditions, respectively, significantly outperforming the next best model, NMO (0.0483 and 0.0483), with an error reduction of about 36\%. On the Navier-Stokes dataset, \method{} achieves an MSE of 0.0714 (regular) and 0.0772 (OOD), lower than NMO's 0.1021 and 0.1032. On the Spherical-SWE dataset, \method{} achieves an MSE of 0.0018 and 0.0020, also outperforming other models like FNO (0.0038 and 0.0045). For the ERA5 dataset, \method{}'s MSE (0.0322 and 0.0321) is better than models like ResNet and FNO, demonstrating \method{}'s superior adaptability and robustness across different conditions.

\begin{table*}[!t]
    \caption{Transfer the model pre-trained from full-data ERA5 dataset to limited-data SEVIR dataset. The results are presented in the formalization of $E\to S$, where $E$ denotes the performance of a model trained from scratch solely on SEVIR, and $S$ represents the performance after fine-tuning with the ERA5 pre-trained model.}
    \vspace{-10pt}
    \label{tab:transfer2}
    \centering
    \begin{small}
            \renewcommand{\multirowsetup}{\centering}
            \resizebox{0.98\textwidth}{!}{%
            \begin{tabular}{l|l|l|l|l|l}
                \toprule
                & \multicolumn{1}{c}{20\% SEVIR} & \multicolumn{1}{c}{40\% SEVIR} & \multicolumn{1}{c}{60\% SEVIR} & \multicolumn{1}{c}{80\% SEVIR} & \multicolumn{1}{c}{100\% SEVIR} \\
                \midrule
                SimVP & 0.37$\to$0.36 (-2.70\%) & 0.36$\to$0.34 (-5.56\%) & 0.29$\to$0.31 (+6.90\%) & 0.25$\to$0.26 (+4.00\%) & 0.19$\to$0.22 (+15.79\%) \\
                \rowcolor{mygrey}SimVP + \method{} & \textbf{0.28$\to$0.26} (-7.14\%) & \textbf{0.27$\to$0.24} (-11.11\%) & \textbf{0.25$\to$0.22} (-12.00\%) & \textbf{0.21$\to$0.19} (-9.52\%) & \textbf{0.18$\to$0.16} (-11.11\%) \\
                \midrule
                PredRNN & 0.62$\to$0.58 (-6.45\%) & 0.52$\to$0.41 (-21.15\%) & 0.42$\to$0.33 (-21.43\%) & 0.27$\to$0.24 (-11.11\%) & 0.23$\to$0.25 (+8.70\%) \\
                \rowcolor{mygrey}PredRNN + \method{} & \textbf{0.30$\to$0.27} (-10.00\%) & \textbf{0.28$\to$0.26} (-7.14\%) & \textbf{0.27$\to$0.24} (-11.11\%) & \textbf{0.25$\to$0.23} (-8.00\%) & \textbf{0.22$\to$0.19} (-13.64\%) \\
                \midrule
                Earthfarseer & 0.26$\to$0.25 (-3.85\%) & 0.24$\to$0.24 (0.00\%) & 0.23$\to$0.21 (-8.70\%) & 0.22$\to$0.19 (-13.64\%) & 0.16$\to$0.17 (+6.25\%) \\
                \rowcolor{mygrey}Earthfarseer + \method{} & \textbf{0.24$\to$0.22} (-8.33\%) & \textbf{0.21$\to$0.18} (-14.29\%) & \textbf{0.19$\to$0.17} (-10.53\%) & \textbf{0.17$\to$0.16} (-5.88\%) & \textbf{0.15$\to$0.13} (-13.33\%) \\
                \bottomrule
            \end{tabular}}
    \end{small}
    \vspace{-8pt}
\end{table*}

\begin{figure*}[t]
\begin{center}
\centerline{\includegraphics[width=0.98\textwidth]{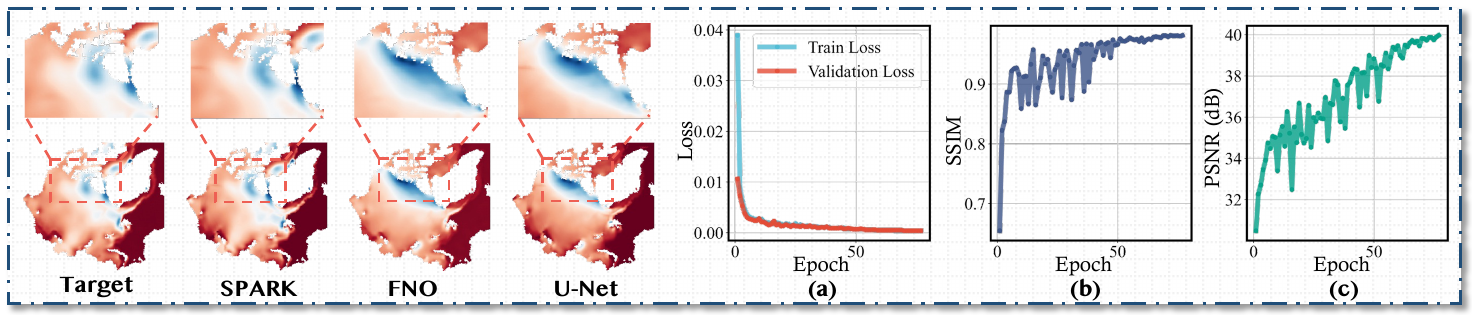}}
\vspace{-10pt}
\caption{The figures in the left show the predictions of different models (\method{}, FNO, and U-Net) for sea ice data at 10-th timestep. The figures in the right demonstrate the curves of loss and performance metrics during training.}
\label{fig:challenging}
\end{center}
\vspace{-8pt}
\end{figure*}

\method{} shows better performance under both in-distribution (ID) and out-of-distribution (OOD) conditions on all benchmark datasets, clearly outperforming existing baseline models. This improvement is most evident in handling complicated fluid dynamics data (e.g., Navier-Stokes and Spherical-SWE) and real meteorological data (e.g., ERA5), where \method{}'s error is significantly lower than other models, proving its strong generalization ability.

Beyond quantitative metrics, the visual results presented in Figure~\ref{fig:vismain} offer qualitative evidence of \method{}'s efficacy. In both the ERA5 and Navier-Stokes datasets, \method{}'s predictions are visually almost indistinguishable from the ground truth. Our model accurately captures high-frequency, complex details that other baselines. Specifically, in Figure~\ref{fig:vismain}(a) (ERA5), FNO and ViT produce overly smoothed predictions, blurring the sharp gradients of temperature fronts and failing to capture the fine-grained structures of wind velocity fields. \method{}, conversely, renders these features with high fidelity. Similarly, in Figure~\ref{fig:vismain}(b) (Navier-Stokes), our model preserves the intricate filament structures and dynamics of small-scale vortices over time, whereas baseline models quickly accumulate errors. 
This indicates that \method{} possesses stronger generalization and robustness in complex spatiotemporal dynamic predictions.

\subsection{Performance Analysis on Challenging Tasks}

To rigorously test our model's capabilities, we evaluate \method{} on the challenging task of \textbf{\textit{sea ice}} forecasting. This problem is notoriously difficult due to the complex, nonlinear dynamics governing sea ice, which involve Lagrangian advection, intricate thermodynamics, and the presence of sharp discontinuities at the ice edge~\citep{notz2012challenges}. Such systems are often a stumbling block for standard data-driven models that tend to produce overly smooth or physically inconsistent predictions.

\noindent\textbf{\textit{Qualitative Analysis.}} The visual results in Figure~\ref{fig:challenging} (left) provide strong evidence of \method{}'s superiority. Our model's predictions are visually almost indistinguishable from the ground truth, accurately preserving the sharp gradients of the ice edge and capturing fine-scale filamentary structures within the ice pack. This is in stark contrast to the baseline models. FNO, for instance, suffers from significant numerical diffusion, resulting in blurry and overly smooth predictions that lose critical high-frequency details. Similarly, U-Net fails to resolve the complex boundaries, indicating a struggle to model the underlying advective-diffusive processes. \method{}'s ability to maintain these sharp features suggests it has learned a more physically faithful representation of the system's dynamics.

\noindent\textit{\textbf{Quantitative Validation.}} The visual fidelity is corroborated by a robust quantitative analysis of the training dynamics (Figure~\ref{fig:challenging}, right). The loss curves (Figure~\ref{fig:challenging}(a)) demonstrate rapid and stable convergence for both training and validation sets, without any signs of overfitting. This efficient learning is mirrored in the performance metrics. Both the SSIM (Figure~\ref{fig:challenging}(b)) and the PSNR (Figure~\ref{fig:challenging}(c)) increase rapidly, saturating at excellent values of approximately 0.95 and 40 dB, respectively. These high scores quantitatively confirm that \method{} achieves a high-fidelity reconstruction that preserves both structural integrity and pixel-level accuracy.

\begin{figure*}[t]
    \centering
    \includegraphics[width=0.99\textwidth]{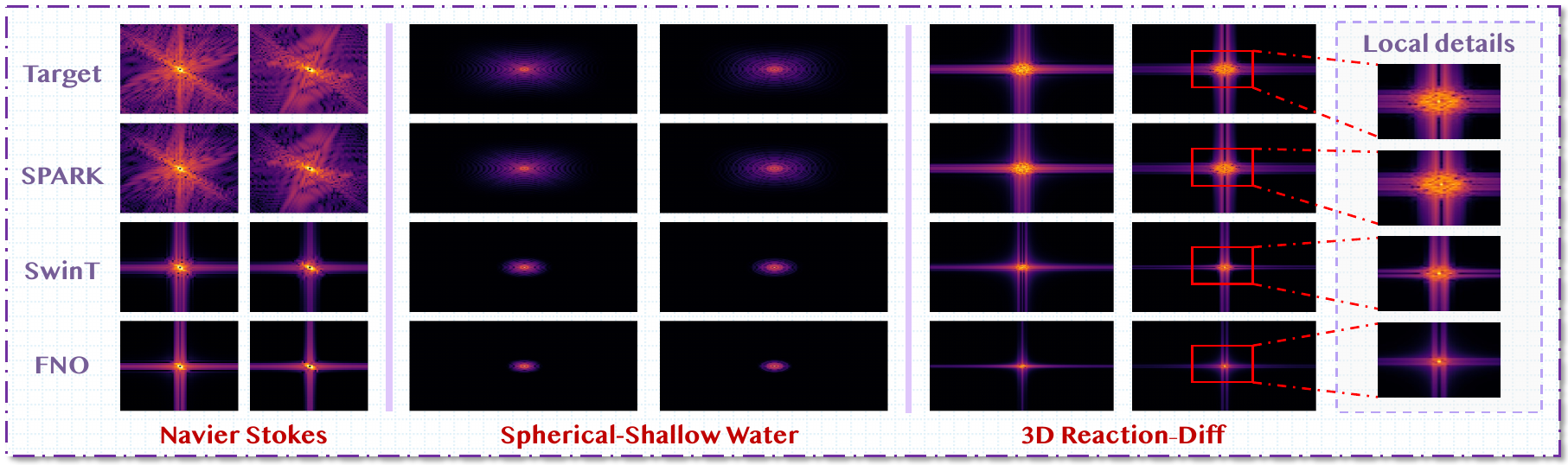}
    \vspace{-5pt}
    \caption{Energy Spectrum Comparison Results on Navier Stokes, Spherical-Shallow Water, and 3D Reaction–Diff datasets.}
    \label{fig:phy} 
\end{figure*}

\subsection{Transferability of SPARK}

In the experimental design for transfer capability, we transfer the pretrained model from the full ERA5 dataset to the data-limited SEVIR dataset to evaluate \method{}'s cross-domain transfer ability. The experiment is based on three baseline models (SimVP~\citep{tan2022simvp}, PredRNN~\citep{wang2017predrnn}, Earthfarseer~\citep{wu2024earthfarsser}), comparing their performance with and without using \method{}. To better demonstrate the effectiveness of transfer learning, we pretrain on the full ERA5 dataset and then fine-tune on different amounts of SEVIR dataset (20\%, 40\%, 60\%, 80\%, 100\%), using mean squared error (MSE x 100) as the evaluation metric. Results as shown in Table~\ref{tab:transfer2}.

For all baseline models, using \method{} for fine-tuning significantly reduces the error on the SEVIR dataset. For instance, Earthfarseer's error decreases from 0.21 to 0.18 (a reduction of 14.29\%) after pretraining on 40\% of ERA5 data, which demonstrate a significant advantage over training without \method{} (error remains unchanged). This indicates that \method{} effectively leverages physics-aware information to enhance the model's transfer learning capability.  

\begin{table}[t]
\centering
\caption{Pre-trained models of different sizes.}
\vspace{-5pt}
\label{tab:sizes}
\resizebox{0.8\columnwidth}{!}{
\begin{tabular}{lcccc}
\toprule
Model Size     & Prometheus  & ERA5   & NS  & 3D-RD     \\ 
\midrule
24.56MB        &  0.0273 & 0.0302 & 0.0723  & 0.0115    \\
17.65MB        &  0.0292 & 0.0323 & 0.0733  & 0.0137    \\ 
9.43MB         &  0.0317 & 0.0342 & 0.0734  & 0.0164    \\ 
4.57MB         &  0.0338 & 0.0388 & 0.0787  & 0.0182    \\ 
2.18MB         &  0.0345 & 0.0391 & 0.0798  & 0.0190    \\
\bottomrule
\end{tabular}
}
\vspace{-5pt}
\end{table}

With smaller data amounts (e.g., 20\% of ERA5 data), models using \method{} show greater performance improvements. For SimVP-V2, when pretrained with 20\% of the data, MSE decreases from 0.28 to 0.26 (a reduction of 7.14\%), while without \method{}, the reduction is only 2.70\%. 
This result demonstrates that \method{} exhibits strong transfer capabilities, particularly in data-scarce scenarios, effectively addressing data insufficiency in cross-domain environments and significantly enhancing model generalization.

\subsection{Analysis of Consistency and Scalability}

In this section, we focus on examining the physical consistency and scalability of the \method{} framework. 
Drawing from the energy spectrum visualization in Figure~\ref{fig:phy} and the performance metrics of pretrained models of different sizes in Table~\ref{tab:sizes}, we have two key observations as follows:

\noindent\textbf{\textit{Consistency Analysis.}}
The Figure~\ref{fig:phy} shows that the energy spectra generated by \method{} are very close to real data, especially in complex fluid dynamics scenarios like Navier-Stokes and Spherical Shallow Water. Compared to other baseline models such as Swin-T and FNO, \method{} better captures the complex details of dynamic systems. This performance is due to \method{} incorporating physical priors, which effectively enhances its ability to model physical phenomena and improves its physical consistency.

\noindent\textbf{\textit{Scalability Analysis.}}
The Table~\ref{tab:sizes} shows that pretrained models of different sizes have similar error rates on the ERA5 and Navier-Stokes datasets. For example, when the model size decreases from 24.56 MB to 9.43 MB, the error on ERA5 only increases slightly from 0.0302 to 0.0342. This indicates that \method{} has good scalability, maintaining high prediction accuracy even when the model is reduced in size. This is due to the physics-aware mechanisms in \method{}. They preserve key physical features, reduce complexity, and ultimately ensure stable model performance.


\subsection{Comparison with OOD-specific Models}

To highlight the capability of our model in handling OOD scenarios, we further select state-of-the-art OOD baselines for comparison. In particular, we include LEADS~\citep{kirchmeyer2022generalizing}, CODA~\citep{yin2021leads} and NUWA~\citep{wang2024nuwadynamics} in our experiments.
These are selected due to their relevance in tackling dynamics modeling and boundary condition incorporation, which aligns with our focus on fluid dynamics modeling under different input conditions. Experimental results on three benchmark datasets are illustrated in Table~\ref{tab:oodmore}.

\begin{table}[t]
\centering
\caption{Performance comparison of different models with and without OOD conditions on the Prometheus, ERA5, and Spherical-SWE benchmark datasets.}
\vspace{-5pt}
\label{tab:oodmore}
\renewcommand{\arraystretch}{1.1}
\resizebox{1.0\columnwidth}{!}{
\begin{tabular}{l|cc|cc|cc}
\toprule
\multirow{4}{*}{Model} & \multicolumn{6}{c}{Benchmarks}  \\
\cmidrule(lr){2-7}
& \multicolumn{2}{c}{Prometheus} & \multicolumn{2}{c}{ERA5}  
& \multicolumn{2}{c}{Spherical-SWE}  \\
\cmidrule(lr){2-7}
& \textit{w/o} OOD & \textit{w/} OOD & \textit{w/o} OOD & \textit{w/} OOD & \textit{w/o} OOD & \textit{w/} OOD   \\
\midrule
LEADS \cite{kirchmeyer2022generalizing} & 0.0374 & 0.0403 & 0.2367 & 0.4233 & 0.0038 & 0.0047  \\
CODA \cite{yin2021leads} & 0.0353 & 0.0372 & 0.1233  & 0.2367  & 0.0034  & 0.0043 \\
NUWA \cite{wang2024nuwadynamics}  & 0.0359 & 0.0398  & 0.0645 & 0.0987 & 0.0032 & 0.0039  \\
\rowcolor{mygrey}\method{} (Ours) & \textbf{0.0323}  & \textbf{0.0328}  & \textbf{0.0398} & \textbf{0.0401}  & \textbf{0.0022} & \textbf{0.0024} \\
\bottomrule
\end{tabular}
\vspace{-5pt}
}
\end{table}

The results show that \method{} performs significantly better in handling OOD scenarios, especially on the Prometheus, ERA5, and Spherical-SWE benchmarks. On the Prometheus dataset, \method{} achieves errors of 0.0323 and 0.0328 for w/o OOD and w/ OOD, respectively, clearly outperforming LEADS (0.0374 and 0.0403), CODA (0.0353 and 0.0372), and NUWA (0.0359 and 0.0398), showing strong generalization and adaptability to changes.
On the ERA5 dataset, \method{}'s errors are 0.0398 and 0.0401, while other models show significantly increased errors under OOD, such as LEADS increasing from 0.2367 to 0.4233, indicating better robustness of \method{} under complex atmospheric conditions. 
In the Spherical-SWE dataset, \method{} achieves errors of 0.0022 and 0.0024 for w/o OOD and w/ OOD, respectively. 
While other models like LEADS show errors of 0.0038 and 0.0047, further proving \method{}'s stability and accuracy in handling boundary condition changes and complex systems.
On the Spherical-SWE dataset, \method{} achieves errors of 0.0022 and 0.0024 for w/o OOD and w/ OOD settings, respectively, compared to errors of 0.0038 and 0.0047 by models like LEADS. This further demonstrates \method{}'s stability and accuracy in handling boundary condition changes and complex spatiotemporal systems.
The results clearly demonstrate that \method{} has excellent performance, high robustness, and strong generalization in OOD scenarios.

\section{Conclusion}
We introduce \method{} for modeling space-time continuous dynamics from non-discretized partial obvervations within complicated physical dynamical systems. 
Our model constructs spatio-temporal graphs to achieve fusion of time and space, and proposes historical continuous dynamics learning within the effectively encoded latent space to obtain the historical spatio-temporal representations.
Ultimately, we designed a fourier-enhanced graph ODE for extrapolation at arbitrary times and locations.
In experiments, \method{} demonstrates state-of-the-art performance of predicting future continuous dynamics on both synthetic and real-world datasets, and its efficient modeling capabilities for partial observations. 

\normalem  
\bibliographystyle{ACM-Reference-Format}
\bibliography{sample-base}

\appendix
\newpage
\appendix

\section{The Proposed \method{} Algorithm}


The complete training procedure for \textsc{SPARK} involves two main steps: state dictionary construction and downstream task training, which we elucidate in two separate algorithms.

\begin{algorithm}[h]
\caption{Pre-training of Physics-Rich State Dictionary}
\label{alg:spark_pretraining}
\begin{algorithmic}[1]
\REQUIRE Training data $\mathcal{D}_{pre} = \{(\mathcal{X}_t, \mathcal{B}_t, \boldsymbol{\lambda}_t)\}$, learning rate $\eta_{pre}$.
\STATE Initialize GNN encoder parameters $\boldsymbol{\Theta}_E$, decoder parameters $\boldsymbol{\Theta}_{\mathcal{I}}$, and state dictionary embeddings $\mathcal{M}$.

\FOR{number of pre-training epochs}
    \STATE Sample a mini-batch $\{(\mathcal{X}, \mathcal{B}, \boldsymbol{\lambda})\}$ from $\mathcal{D}_{pre}$.
    
    \STATE \texttt{// Stage 1: Physics-Informed Feature Encoding}
    \STATE Fused features $\boldsymbol{h} \leftarrow \text{EncodePhysicalPriors}(\mathcal{X}, \mathcal{B}, \boldsymbol{\lambda}; \boldsymbol{\Theta}_E)$.
    
    \STATE \texttt{// Stage 2: Discretize via Vector Quantization}
    \STATE Discretized $\boldsymbol{z} \leftarrow \text{VectorQuantization}(\boldsymbol{h}, \mathcal{M})$.
    
    \STATE \texttt{// Stage 3: Reconstruction}
    \STATE Reconstructed observations $\hat{\mathcal{X}} \leftarrow \text{Decoder}(\boldsymbol{z}; \boldsymbol{\Theta}_{\mathcal{I}})$.
    
    \STATE \texttt{// Optimization}
    \STATE Compute pre-training loss $\mathcal{L}_{pre}(\mathcal{X}, \hat{\mathcal{X}}, \boldsymbol{h}, \boldsymbol{z})$. \COMMENT{Combines reconstruction and commitment loss}
    \STATE Update parameters by descending the gradient: $\{\boldsymbol{\Theta}_E, \boldsymbol{\Theta}_{\mathcal{I}}, \mathcal{M}\} \leftarrow \{\dots\} - \eta_{pre} \nabla \mathcal{L}_{pre}$.
\ENDFOR

\RETURN Trained Encoder $\boldsymbol{\Theta}_E^*$ and State Dictionary $\mathcal{M}^*$.
\end{algorithmic}
\end{algorithm}

\begin{algorithm}[h]
\caption{Downstream Training with \textsc{SPARK} Augmentation}
\label{alg:spark_downstream}
\begin{algorithmic}[1]
\REQUIRE Training data $\mathcal{D}_{dyn} = \{(\mathcal{X}_t, \mathcal{Y}_t, \mathcal{B}_t, \boldsymbol{\lambda}_t)\}$, learning rate $\eta_{dyn}$.
\REQUIRE Pre-trained Encoder $\boldsymbol{\Theta}_E^*$ and State Dictionary $\mathcal{M}^*$ (from Algorithm~\ref{alg:spark_pretraining}).
\STATE Initialize Fourier-enhanced Graph ODE parameters $\boldsymbol{\Theta}_{\Phi}$.
\STATE Freeze parameters of the pre-trained encoder $\boldsymbol{\Theta}_E^*$ and state dictionary $\mathcal{M}^*$.

\FOR{number of training epochs}
    \STATE Sample a mini-batch $\{(\mathcal{X}, \mathcal{Y}, \mathcal{B}, \boldsymbol{\lambda})\}$ from $\mathcal{D}_{dyn}$.
    
    \STATE \texttt{// Stage 1: Memory-Bank Guided Data Augmentation (on-the-fly)}
    \STATE Initial features $\boldsymbol{h} \leftarrow \text{EncodePhysicalPriors}(\mathcal{X}, \mathcal{B}, \boldsymbol{\lambda}; \boldsymbol{\Theta}_E^*)$.
    \STATE Augmented features $\boldsymbol{h}_{aug} \leftarrow \text{Augment}(\boldsymbol{h}, \mathcal{M}^*)$. \COMMENT{Top-K search and interpolation}
    
    \STATE \texttt{// Stage 2: Dynamics Prediction}
    \STATE Predicted future states $\hat{\mathcal{Y}} \leftarrow \text{FourierGraphODE}(\boldsymbol{h}_{aug}; \boldsymbol{\Theta}_{\Phi})$.
    
    \STATE \texttt{// Optimization}
    \STATE Compute dynamics prediction loss $\mathcal{L}_{dyn}(\hat{\mathcal{Y}}, \mathcal{Y})$.
    \STATE Update parameters by descending the gradient: $\boldsymbol{\Theta}_{\Phi} \leftarrow \boldsymbol{\Theta}_{\Phi} - \eta_{dyn} \nabla_{\boldsymbol{\Theta}_{\Phi}} \mathcal{L}_{dyn}$.
\ENDFOR

\RETURN Trained dynamics model parameters $\boldsymbol{\Theta}_{\Phi}^*$.
\end{algorithmic}
\end{algorithm}

\textbf{Algorithm~\ref{alg:spark_pretraining} describes the pre-training stage,} where the core task is to learn how to encode physical priors into a discrete state dictionary for later augmentation. 
Subsequently, \textbf{Algorithm~\ref{alg:spark_downstream} details the training of the dynamics prediction model,} where the pre-trained modules are used to perform on-the-fly augmentation, thereby enhancing the model's generalization capabilities.

\section{Proofs of Theorem~\ref{theorem1}}\label{sec:theo1}


We prove that introducing physical priors improves model generalization from an information-theoretic perspective, based on the relationship between mutual information and generalization error.

\subsection{Preliminaries}

\subsubsection{Generalization Error and Empirical Error} ~

\noindent $\triangleright$ True Risk (Expected Loss):
\begin{equation}
\mathcal{L}(\theta) = \mathbb{E}_{(X, Y) \sim \mathcal{D}_{\text{true}}}[\ell(\theta; X, Y)],
\end{equation}
where $\ell(\theta; X, Y)$ is the loss function, and $\mathcal{D}_{\text{true}}$ is the true data distribution.

\noindent $\triangleright$ Empirical Risk:
\begin{equation}
\mathcal{L}_{\text{emp}}(\theta; \mathcal{D}) = \dfrac{1}{n} \sum_{i=1}^n \ell(\theta; x_i, y_i),
\end{equation}
where $\mathcal{D} = \{(x_i, y_i)\}_{i=1}^n$ is the training dataset.

\subsubsection{Mutual Information}~

\noindent $\triangleright$ Mutual Information:
\begin{equation}
I(\theta; \mathcal{D}) = \mathbb{E}_{\theta, \mathcal{D}} \left[ \log \dfrac{p(\theta, \mathcal{D})}{p(\theta) p(\mathcal{D})} \right].
\end{equation}
\noindent $\triangleright$ Conditional Mutual Information (given physical prior $\mathcal{P}$):
\begin{equation}
I(\theta; \mathcal{D} \mid \mathcal{P}) = \mathbb{E}_{\theta, \mathcal{D}, \mathcal{P}} \left[ \log \dfrac{p(\theta, \mathcal{D} \mid \mathcal{P})}{p(\theta \mid \mathcal{P}) p(\mathcal{D} \mid \mathcal{P})} \right].
\end{equation}

\subsection{Proof}

We then prove that introducing physical prior information reduces the upper bound of the generalization error.


\noindent\textbf{Relating Generalization Error to Mutual Information.}
According to information-theoretic results\footnote{Reference: Xu, A., \& Raginsky, M. (2017). Information-theoretic analysis of generalization capability of learning algorithms. \textit{Advances in Neural Information Processing Systems}, 30.}, for any learning algorithm, the expected generalization error has the following upper bound:
\begin{equation}
\left| \mathbb{E}_{\theta, \mathcal{D}} \left[ \mathcal{L}(\theta) - \mathcal{L}_{\text{emp}}(\theta; \mathcal{D}) \right] \right| \leq \sqrt{\dfrac{2 I(\theta; \mathcal{D})}{n}}.
\end{equation}
\noindent\textbf{Introducing Physical Prior Information.}
When we introduce physical prior information $\mathcal{P}$, we consider the conditional mutual information $I(\theta; \mathcal{D} \mid \mathcal{P})$. Since $\mathcal{P}$ is known, we can reconsider the upper bound on the generalization error under the condition of $\mathcal{P}$.

\noindent\textbf{Recalculate the Upper Bound of Generalization Error.}
Based on conditional mutual information, the upper bound becomes:
\begin{equation}
\left| \mathbb{E}_{\theta, \mathcal{D}, \mathcal{P}} \left[ \mathcal{L}(\theta) - \mathcal{L}_{\text{emp}}(\theta; \mathcal{D}) \mid \mathcal{P} \right] \right| \leq \sqrt{\dfrac{2 I(\theta; \mathcal{D} \mid \mathcal{P})}{n}}.
\end{equation}
Since $\mathcal{P}$ is fixed, we can take the expectation over $\mathcal{P}$:
\begin{equation}
\left| \mathbb{E}_{\theta, \mathcal{D}} \left[ \mathcal{L}(\theta) - \mathcal{L}_{\text{emp}}(\theta; \mathcal{D}) \right] \right| \leq \sqrt{\dfrac{2 I(\theta; \mathcal{D} \mid \mathcal{P})}{n}}.
\end{equation}
\noindent\textbf{Physical Prior Reduces Mutual Information.}
The physical prior $\mathcal{P}$ provides additional knowledge about the model parameters $\theta$, which reduces the mutual information $I(\theta; \mathcal{D} \mid \mathcal{P})$ between $\theta$ and $\mathcal{D}$ under the condition $\mathcal{P}$.

Intuitively, the physical prior restricts the possible parameter space, reducing the influence of training data on parameters and thus decreasing the model's dependence on the training data. Mathematically, mutual information satisfies:
\begin{equation}
I(\theta; \mathcal{D}) \geq I(\theta; \mathcal{D} \mid \mathcal{P}).
\end{equation}
Combining the above steps, we conclude:
\begin{equation}
\left| \mathbb{E}_{\theta, \mathcal{D}} \left[ \mathcal{L}(\theta) - \mathcal{L}_{\text{emp}}(\theta; \mathcal{D}) \right] \right| \leq \sqrt{\dfrac{2 I(\theta; \mathcal{D} \mid \mathcal{P})}{n}} \leq \sqrt{\dfrac{2 I(\theta; \mathcal{D})}{n}}.
\end{equation}
Thus, introducing physical prior information $\mathcal{P}$ reduces the mutual information $I(\theta; \mathcal{D} \mid \mathcal{P})$, leading to a reduced upper bound on generalization error, and thereby improving the model's generalization capability.

\subsection{Generalization Error Bound in Bayesian Learning with Physical Prior}

Consider a hypothesis space $\mathcal{H}$, with model parameters $\theta \in \mathcal{H}$, a training dataset $\mathcal{D} = \{(x_i, y_i)\}_{i=1}^n$, and a loss function $\ell(\theta; x, y)$. The true risk is:
\begin{equation}
\mathcal{L}(\theta) = \mathbb{E}_{(x, y) \sim \mathcal{P}_{\text{data}}}[\ell(\theta; x, y)],
\end{equation}
and the empirical risk is:
\begin{equation}
\mathcal{L}_{\text{emp}}(\theta) = \frac{1}{n} \sum_{i=1}^n \ell(\theta; x_i, y_i).
\end{equation}
Assume the prior distribution $P(\theta)$ contains physical prior information, and the posterior distribution is $Q(\theta)$. For any $\delta > 0$, with probability at least $1 - \delta$, the generalization error has the following upper bound:
\begin{equation}
\mathbb{E}_{\theta \sim Q} [\mathcal{L}(\theta)] \leq \mathbb{E}_{\theta \sim Q} [\mathcal{L}_{\text{emp}}(\theta)] + \sqrt{\dfrac{KL(Q \| P) + \ln \dfrac{2\sqrt{n}}{\delta}}{2n}},
\end{equation}
where $KL(Q \| P)$ is the Kullback-Leibler divergence between the posterior $Q$ and the prior $P$.


Introducing physical prior information as the prior distribution $P(\theta)$ reduces the Kullback-Leibler divergence $KL(Q \| P)$ between the posterior $Q(\theta)$ and the prior $P(\theta)$. According to the theorem, this reduces the upper bound of the generalization error. This implies that, within the Bayesian learning framework, incorporating physical prior information enhances the model's generalization ability, leading to better performance on unseen data.


\section{Proofs of Theorem~\ref{theorem2}}\label{sec:theo2}


We prove that introducing physical priors improves model generalization from a Bayesian learning perspective, using the PAC-Bayesian theory. The PAC-Bayesian theorem provides an upper bound on the generalization error for randomized algorithms, which relates to the KL divergence between prior and posterior distributions.

\subsection{Preliminaries}

\noindent$\triangleright$ \textbf{PAC-Bayesian Theorem.}
The PAC-Bayesian theorem provides a probabilistic upper bound on the generalization performance of randomized learning algorithms. The core idea is to define a probability distribution over the hypothesis space and use the KL divergence between the prior and posterior to quantify generalization error.

\noindent$\triangleright$ \textbf{KL Divergence (Relative Entropy).}
For two probability distributions $P$ and $Q$, the KL divergence is defined as:
\begin{equation}
KL(Q \| P) = \int \ln \left( \dfrac{dQ}{dP} \right) dQ.
\end{equation}
The KL divergence measures how far the distribution $Q$ deviates from $P$.

\subsection{Proof}


\noindent\textbf{Define the Randomized Prediction Function.}
In the Bayesian framework, the model parameter $\theta$ is treated as a random variable, whose distribution is given by the posterior distribution $Q(\theta)$. During prediction, the model samples $\theta$ from the posterior and uses it for prediction.

\noindent\textbf{Introduce Physical Prior Information.}
Physical prior information is encoded in the prior distribution $P(\theta)$. This prior reflects our belief about the model parameters before observing data.

\noindent\textbf{Apply the PAC-Bayesian Theorem.}
According to the PAC-Bayesian theorem, for any posterior distribution $Q(\theta)$, with probability at least $1 - \delta$, we have:
\begin{equation}
\mathbb{E}_{\theta \sim Q} [\mathcal{L}(\theta)] \leq \mathbb{E}_{\theta \sim Q} [\mathcal{L}_{\text{emp}}(\theta)] + \sqrt{\dfrac{KL(Q \| P) + \ln \dfrac{2\sqrt{n}}{\delta}}{2n}}.
\end{equation}
\textbf{Note}: The full proof of this theorem involves the Hoeffding inequality and variations of martingale inequalities, but here we focus on applying the conclusion.

\noindent\textbf{Interpret the Role of KL Divergence.}
The smaller the KL divergence $KL(Q \| P)$, the closer the posterior distribution $Q$ is to the prior distribution $P$. This means that the model deviates less from the prior information during learning.
Introducing physical prior information makes the prior distribution $P(\theta)$ closer to the true parameter distribution, reducing the KL divergence between the posterior $Q(\theta)$ and prior $P(\theta)$, i.e., $KL(Q \| P)$ decreases.

\noindent\textbf{Derive the Reduction in Generalization Error Bound.}
Since $KL(Q \| P)$ decreases, the PAC-Bayesian upper bound on the generalization error also decreases:
\begin{equation}
\mathbb{E}_{\theta \sim Q} [\mathcal{L}(\theta)] - \mathbb{E}_{\theta \sim Q} [\mathcal{L}_{\text{emp}}(\theta)] \leq \sqrt{\dfrac{\downarrow KL(Q \| P) + \ln \dfrac{2\sqrt{n}}{\delta}}{2n}}.
\end{equation}
Thus, introducing physical prior information reduces the upper bound on the generalization error, thereby improving the model's generalization capability.


\subsection{Discussion}

\subsubsection{Role of Physical Prior Information}

$\triangleright$ \textbf{Narrowing the Parameter Space}: Physical priors restrict the possible values of model parameters, making the prior distribution $P(\theta)$ more concentrated in regions that follow physical laws.
$\triangleright$ \textbf{Guiding the Posterior Distribution}: Since the prior distribution includes physical information, the posterior distribution tends to favor parameter regions that are consistent with physical laws during the update.

\subsubsection{Relationship between KL Divergence and Generalization Error}

$\triangleright$ \textbf{KL Divergence as a Measure of Deviation}: KL divergence measures how much the posterior deviates from the prior. The smaller the deviation, the lower the upper bound on the generalization error.
$\triangleright$ \textbf{Coordination between Prior and Posterior}: A good prior allows the model to make less drastic adjustments to parameters given the data, thereby maintaining model stability and generalizability.

\subsubsection{Advantages of the Bayesian Learning Framework}

$\triangleright$ \textbf{Naturally Incorporates Prior Knowledge}: The Bayesian approach allows prior knowledge to be incorporated into the model as a probability distribution, which helps improve model performance, especially when data is limited.
$\triangleright$ \textbf{Probabilistic Interpretation}: The PAC-Bayesian theorem provides an upper bound on the generalization error with probabilistic guarantees, making the theoretical results more robust.

\section{Detailed Description of Datasets} \label{datadescrip}

We evaluate our proposed \method{} on benchmark datasets in three fields:
\textit{Prometheus} for computational fluid dynamics;
\textit{ERA5} for real-world scenarios;
\textit{2D Navier-Stokes Equations}, \textit{Spherical Shallow Water Equations}, and \textit{3D Reaction-Diffusion Equations} for partial differential equations.

\textbf{\textit{Prometheus}}~\citep{wu2024prometheus}
is a large-scale, out-of-distribution (OOD) fluid dynamics dataset designed for the development and benchmarking of machine learning models, particularly those that predict fluid dynamics under varying environmental conditions. 
This dataset includes simulations of tunnel and pool fires (representated as Prometheus-T and Prometheus-P in experiments), encompassing a wide range of fire dynamics scenarios modeled using fire dynamics simulators that solve the Navier-Stokes equations. Key features of the dataset include 25 different environmental settings with variations in parameters such as Heat Release Rate (HRR) and ventilation speeds.
In total, the Prometheus dataset encompasses 4.8 TB of raw data, which is compressed to 340 GB. 
It not only enhances the research on fluid dynamics modeling but also aids in the development of models capable of handling complex, real-world scenarios in safety-critical applications like fire safety management and emergency response planning.

\textbf{\textit{ERA5}}~\citep{hersbach2020era5}
is a global atmospheric reanalysis dataset developed by the European Centre for Medium-Range Weather Forecasts (ECMWF), offering comprehensive weather data from 1979 to the present with exceptional spatial resolution (31 km) and hourly temporal granularity. This dataset encompasses a rich array of meteorological variables, including but not limited to surface pressure, sea surface temperature, sea surface height, and two-meter air temperature. \textit{ERA5} is extensively employed across a multitude of domains, including climate modeling, environmental monitoring, atmospheric dynamics research, and energy management optimization. The dataset's integration of physical models with vast observational data makes it a cornerstone for advancing predictive models in meteorology and climate science.

\textbf{\textit{2D Navier-Stokes Equations}}~\citep{li2021fourier}
describe the motion of fluid substances such as liquids and gases. These equations are a set of partial differential equations that predict weather, ocean currents, water flow in a pipe, and air flow around a wing, among other phenomena. The equations arise from applying Newton's second law to fluid motion, together with the assumption that the fluid stress is the sum of a diffusing viscous term proportional to the gradient of velocity, and a pressure term. 

\textbf{\textit{Spherical Shallow Water Equations}}~\citep{galewsky2004initial}
model surface water flows under the assumption of a shallow depth compared to horizontal dimensions. This simplification leads to the Shallow Water equations, a set of partial differential equations (PDEs) that describe the flow below a pressure surface in a fluid (often water).
Shallow Water typically encompasses variables such as water surface elevation and the two components of velocity field ($u$-velocity in the $x$-direction and $v$-velocity in the $y$-direction). These properties are crucial for modeling waves, tides, and large-scale circulations in oceans and atmospheres. The equations consist of a continuity equation for mass conservation and a momentum equation for momentum conservation:
\begin{equation}
\begin{gathered}
    \frac{\partial h}{\partial t} + \nabla \cdot (h \cdot u) = 0, \\
    \frac{\partial u}{\partial t} + (u \cdot \nabla) u + g \nabla h = 0,
\end{gathered}
\end{equation}
where $h$ represents the fluid depth, $u$ is the velocity field, and $g$ denotes the acceleration due to gravity. These equations are used extensively in environmental modeling, including weather forecasting, oceanography, and climate studies.

\textbf{\textit{3D Reaction-Diffusion Equations}}~\citep{rao2023encoding}
are a class of partial differential equations (PDEs) that describe the temporal and spatial evolution of chemical species in three-dimensional domains. These equations are fundamental for modeling systems where chemical substances not only react but also diffuse through a 3D medium. The interaction between reaction kinetics and diffusion mechanisms leads to intricate spatiotemporal dynamics that are critical in fields such as biology, chemistry, and physics. The general form of the 3D reaction-diffusion system can be expressed as:
\begin{equation}
\begin{aligned}
\frac{\partial u}{\partial t} &= D_u \nabla^2 u + f(u, v, w) - g(u, v)u + \alpha_u + \sigma_u S_u(x, y, z, t), \\
\frac{\partial v}{\partial t} &= D_v \nabla^2 v + h(u, v) u^2 - \beta v + \alpha_v + \sigma_v S_v(x, y, z, t), \\
\frac{\partial w}{\partial t} &= D_w \nabla^2 w + p(v, w) - \gamma w + \alpha_w + \sigma_w S_w(x, y, z, t),
\end{aligned}
\end{equation}
where \( u \), \( v \), and \( w \) represent the concentrations of different chemical species, \( D_u \), \( D_v \), and \( D_w \) denote their respective diffusion coefficients. The terms \( f(u, v, w) \) and \( p(v, w) \) describe the reaction kinetics that govern the interactions between these species, while \( g(u, v) \) and \( h(u, v) \) control the rate of conversion and interaction. \( \alpha_u \), \( \alpha_v \), and \( \alpha_w \) represent constant growth rates, and \( \sigma_u \), \( \sigma_v \), and \( \sigma_w \) introduce noise terms that model stochastic external influences through spatially and temporally dependent source functions \( S_u(x, y, z, t) \), \( S_v(x, y, z, t) \), and \( S_w(x, y, z, t) \). The resulting system of equations provides a robust framework for simulating complex dynamical behaviors in three-dimensional reactive-diffusive environments.

\section{External Experiments}

\subsection{Parameter Sensitivity Analysis}

To investigate the influence of hyperparameter $k$, we add experiments on the value of $k$ on the Navier-Stokes, Prometheus, 3D Reaction–Diff, and ERA5 datasets. The candidate values are $\{1,3,5,7,9,11\}$, and the results are shown in Table~\ref{tab:hyper}. 
As $k$ increases, the model's performance first improves and then declines, with optimal performance generally achieved when $k$ is between 3 and 5.

\begin{table}[h!]
\caption{Performance comparison of different $k$.}
\vspace{-5pt}
\label{tab:hyper}
\centering
\resizebox{0.99\linewidth}{!}{
\begin{tabular}{lcccc}
\toprule
\textbf{$k$} & \textbf{Navier–Stokes} & \textbf{Spherical-SWE} & \textbf{Prometheus} & \textbf{3D Reaction–Diff} \\
\midrule
1  & 0.0752 & 0.0022       & 0.0315       & 0.0116       \\
3  & \underline{0.0726} & \textbf{0.0018} & \textbf{0.0296} &  \underline{0.0108}     \\
5  & \textbf{0.0715} & \underline{0.0021}  & \underline{0.0303}  & \textbf{0.0104} \\
7  & 0.0731 & 0.0024       & 0.0311       & 0.0110       \\
9  & 0.0764 & 0.0025       & 0.0320       & 0.0121       \\
11 & 0.0780 & 0.0029       & 0.0327       & 0.0128       \\
\bottomrule
\end{tabular}
}
\end{table}

\subsection{Low-data Regime Experiments}

To explore the performance of our model in very low-data regime of transfer learning, we conduct experiments here.
Specifically, after pre-training on the full ERA5 dataset, we finetune on subsets of the Sevir dataset with varying amounts of data (1\%, 3\%, 5\%, and 10\%). The detailed comparison of baseline models (PredRNN and SimVP) with and without the SPARK plugin is shown in Table~\ref{tab:lowdata}.

\begin{table}[h!]
\caption{Performance comparison of varying data amounts.}
\vspace{-5pt}
\label{tab:lowdata}
\centering
\begin{scriptsize}
\resizebox{1.0\linewidth}{!}{
\begin{tabular}{lcccc}
\toprule
 & \textbf{1\% Sevir} & \textbf{3\% Sevir} & \textbf{5\% Sevir} & \textbf{10\% Sevir} \\
\midrule
PredRNN       & 3.51→3.38  & 2.57→2.35 & 1.83→1.68 & 1.22→1.16  \\
PredRNN+SPARK & 3.37→3.02  & 2.49→2.14 & 1.72→1.45 & 1.14→0.97  \\
SimVP         & 2.43→2.20  & 1.86→1.55 & 1.29→1.11 & 0.75→0.68  \\
SimVP+SPARK   & 2.30→1.98  & 1.75→1.23 & 1.21→0.98 & 0.71→0.57  \\
\bottomrule
\end{tabular}
}
\end{scriptsize}
\end{table}

The results show that models with SPARK plugin consistently outperform their baseline models in the very low data regime.

\subsection{Extreme Event Prediction.}

For extreme event prediction, we use Sevir dataset, which contains data related to severe weather phenomena. To better evaluate the prediction performance of extreme events, we used the Critical Success Index (CSI), in addition to MSE. For simplicity, we used only the thresholds {16, 133, 181, 219} and the mean CSI-M. The results in Table~\ref{tab:extreme} show that our model consistently outperform these baselines in extreme event prediction. 

\begin{table}[h!]
\caption{Extreme event prediction performance.}
\vspace{-5pt}
\label{tab:extreme}
\centering
\resizebox{1.0\linewidth}{!}{
\begin{tabular}{lcccccc}
\toprule
\textbf{Model} & \textbf{CSI-M} $\uparrow$ & \textbf{CSI-219} $\uparrow$ & \textbf{CSI-181} $\uparrow$ & \textbf{CSI-133} $\uparrow$ & \textbf{CSI-16} $\uparrow$ & \textbf{MSE ($10^{-3}$)} $\downarrow$ \\
\midrule
U-Net   & 0.3593 & 0.0577 & 0.1580 & 0.3274 & 0.7441 & 4.1119 \\
ViT     & 0.3692 & 0.0965 & 0.1892 & 0.3465 & 0.7326 & 4.1661 \\
PredRNN & 0.4028 & 0.1274 & 0.2324 & 0.3858 & 0.7507 & 3.9014 \\
SimVP   & 0.4275 & 0.1492 & 0.2538 & 0.4084 & 0.7566 & 3.8182 \\
Ours    & \textbf{0.4683} & \textbf{0.1721} & \textbf{0.2734} & \textbf{0.4375} & \textbf{0.7792} & \textbf{3.6537} \\
\bottomrule
\end{tabular}
}
\end{table}

\section{Details of Model Architecture}

Here, we provided detailed model information using the Navier-Stokes dataset as an example, as shown in Table~\ref{tab:Upstream} and Table~\ref{tab:Downstream}.


\begin{table}[h]
\caption{Details of SPARK's upstream architecture.}
\vspace{-5pt}
\label{tab:Upstream}
\centering
\resizebox{1.0\linewidth}{!}{
\begin{tabular}{llc}
\toprule
\multicolumn{3}{c}{\textbf{Upstream}} \\
\midrule
 \textbf{Procedure} & \textbf{Layer} & \textbf{Dimension} \\
\midrule
Physical parameters injection & Channel attention & (2, 128) \\
 & Aggregation & (4096, 128) \\
GNN reconstruction & Graph Encoder (GNN Layer × L) & (4096, 128) \\
 & BatchNorm + ReLU & (4096, 128) \\
State dictionary & Construction & ($M$, 128) \\
 & Linear + LayerNorm & (4096, 128) \\
\bottomrule
\end{tabular}
}
\end{table}

\begin{table}[h]
\caption{Details of SPARK's downstream architecture.}
\vspace{-5pt}
\label{tab:Downstream}
\centering
\resizebox{0.95\linewidth}{!}{
\begin{tabular}{llc}
\toprule
\multicolumn{3}{c}{\textbf{Downstream}} \\
\midrule
 \textbf{Procedure} & \textbf{Layer} & \textbf{Dimension} \\
\midrule
Augmentation & GNN Encoder & ($T_0$, 4096, 128) \\
 & State dictionary retrieval & ($T_0$, 4096, 128) \\
Historical observations encoding & Attention score of time steps & (1, $T_0$) \\
 & Initial state encoding & (1, 4096, 128) \\
Fourier-enhanced graph ODE & Fourier transform & (1, 4096, 128) \\
 & Linear transform & (1, 4096, 128) \\
 & Inverse Fourier transform & (1, 4096, 128) \\
 & ODE solver & ($T$, 4096, 128) \\
\bottomrule
\end{tabular}
}
\end{table}

\end{document}